\documentclass[conference]{IEEEtran}
\IEEEoverridecommandlockouts
\usepackage{cite}
\usepackage{amsmath,amssymb,amsfonts}
\usepackage{algorithmic}
\usepackage{graphicx}
\usepackage{textcomp}
\usepackage{xcolor}
\def\BibTeX{{\rm B\kern-.05em{\sc i\kern-.025em b}\kern-.08em
    T\kern-.1667em\lower.7ex\hbox{E}\kern-.125emX}}
\usepackage{url}
\usepackage{booktabs}
\newlength\lengtha 
\newlength\lengthb 
\newlength\lengthc 
\newlength\lengthd 
\newcommand{\tabincell}[2]{\begin{tabular}{@{}#1@{}}#2\end{tabular}}
\newcommand{\tablefont}{\fontsize{8pt}{\baselineskip}\selectfont}

\usepackage{tikz}
\usepackage{textcomp}
\usepackage{hyperref}
\usepackage{lipsum}
\newcommand\copyrighttext{%
  \footnotesize \textcopyright 2021 IEEE Copyright.
  }
\newcommand\copyrightnotice{%
\begin{tikzpicture}[remember picture,overlay]
\node[anchor=south,yshift=10pt] at (current page.south) {\fbox{\parbox{\dimexpr\textwidth-\fboxsep-\fboxrule\relax}{\copyrighttext}}};
\end{tikzpicture}%
}

\begin{document}

\title{DeepfakeUCL: Deepfake Detection via Unsupervised Contrastive Learning}

\author{\IEEEauthorblockN{1\textsuperscript{st} Sheldon Fung}
\IEEEauthorblockA{\textit{School of Information Technology} \\
\textit{Deakin University}\\
Geelong, Australia \\
sheldonvon@outlook.com}
\and
\IEEEauthorblockN{2\textsuperscript{nd} Xuequan Lu*
\thanks{{\footnotesize \textsuperscript{*}}Corresponding author.}}
\IEEEauthorblockA{\textit{School of Information Technology} \\
\textit{Deakin University}\\
Geelong, Australia \\
xuequan.lu@deakin.edu.au}
\and
\IEEEauthorblockN{3\textsuperscript{rd} Chao Zhang}
\IEEEauthorblockA{\textit{Faculty of Engineering} \\
\textit{University of Fukui}\\
Fukui, Japan \\
zhang@u-fukui.ac.jp}
\and
\IEEEauthorblockN{4\textsuperscript{th} Chang-Tsun Li}
\IEEEauthorblockA{\textit{School of Information Technology} \\
\textit{Deakin University}\\
Geelong, Australia \\
changtsun.li@deakin.edu.au}
}

\maketitle
\copyrightnotice
\begin{abstract}
Face deepfake detection has seen impressive results recently. Nearly all existing deep learning techniques for face deepfake detection are fully supervised and require labels during training. In this paper, we design a novel deepfake detection method via unsupervised contrastive learning. We first generate two different transformed versions of an image and feed them into two sequential sub-networks, i.e., an encoder and a projection head. The unsupervised training is achieved by maximizing the correspondence degree of the outputs of the projection head. To evaluate the detection performance of our unsupervised method, we further use the unsupervised features to train an efficient linear classification network. Extensive experiments show that our unsupervised learning method enables comparable detection performance to state-of-the-art supervised techniques, in both the intra- and inter-dataset settings. We also conduct ablation studies for our method.

\end{abstract}

\section{Introduction}
Realistic face synthesis or manipulation has led to a rapid increase of deepfake images and videos which pose security and privacy threats to our society. In light of this, researchers have proposed countermeasures against face deepfake, i.e., the detection or recognition of face deepfake. Note that there are different deepfakes, and in this work, we simply use deepfake for face deepfake unless stated otherwise.

It is challenging to detect manipulated faces due to the evolution of deep learning techniques for face manipulation. Early deepfake detection methods focused on the use of cues left by the face manipulation techniques, for example, Li et al. \cite{Blinking2018} determined whether or not an image is manipulated by analyzing the frequency of eye blinking. These methods are fragile and tend to fail if those cues are removed or missing. Deep learning methods were therefore proposed to detect face manipulation, and have achieved promising outcomes due to supervised learning. With technical evolution, those deepfake contents are becoming increasingly realistic and the well-trained models will thus become obsolete and require retraining on new data. Nevertheless, supervised learning has to ``see'' the labels, and labeling is time-consuming and tedious. To our knowledge, unsupervised learning for deepfake detection has rarely been studied so far. It is more challenging than supervised learning since labels are not known during training.

In this paper, we design a novel unsupervised learning approach for face manipulation detection. The core idea is to firstly generate two transformed versions of a face image using two different transformations, and then maximize their agreement after going through an encoder network and a projection head network. This is inspired by a contrastive framework \cite{simCLR2020}. The model trained without supervision will be used to produce features (i.e., output of the encoder) to be taken as the input of a linear classifier network for deepfake evaluation. We use the output of the encoder since it has more effective features than the output of the projection head (see Section \ref{Sec:ablation}).

Extensive experiments on three publicly available datasets validate our unsupervised contrastive learning method. It yields comparable performance to state-of-the-art supervised learning methods and non-deep-learning methods in terms of both the intra-dataset and inter-dataset settings. We also perform ablation studies for our method. Our main contributions are as follows.
\begin{itemize}
    \item We propose an unsupervised contrastive learning approach for deepfake detection.
    \item We conduct a variety of experiments to test our method and compare it with state-of-the-art deepfake detection techniques.
\end{itemize}

The rest of the paper is organized as follows. Section \ref{sec:relatedwork} reviews previous research work on face manipulation and deepfake detection. Section \ref{sec:method} presents the proposed approach. We explain the experimental results and analyze the results in Section \ref{sec:results}. Section \ref{sec:conclusion} concludes this work.

\section{Related Work}
\label{sec:relatedwork}

In this section, we will firstly review previous face synthesis or manipulation methods. Then we will look back upon recent works on the detection of face forgery.

\textbf{Face manipulation approaches.} There have been methods for face image manipulation before the emergence of deep learning based methods. Dale et al. \cite{Dale2011} introduced a face-swapping method based on a 3D multi-linear model for face tracking and warping. A similar approach was presented by Justus et al. \cite{Justus2016}, which tracks the facial expression using a dense photometric consistency measure. Zhmoginov et al. \cite{Zhmoginov2016} introduced deep learning techniques into the task of face manipulation. Concretely, they used neural networks to effectively invert low-dimensional face embeddings while producing realistically looking consistent images, which was later further implemented into a famous phone app called FaceAPP \cite{Faceapp2020}. Since then, many deep learning based face manipulation methods arise. For example, Thies et al. \cite{Thies2019} introduced a new image synthesis approach that takes advantage of the traditional graphics pipeline and learnable components, which was later used to generate manipulated faces in FaceForensics++ \cite{Andreas2019}. An even more intriguing and similar technique was introduced by Wei et al. \cite{Wei2017}, which aimed at modifying a face image according to a given attribute value. Shao et al. \cite{shao2019explicit} proposed to explicitly transfer expressions by directly mapping two unpaired images to two synthesized images with swapped expressions.

\textbf{Detection methods.} The privacy and security impact of face manipulation on individuals as well as society drives researchers to develop detection techniques for face manipulation. Methods for detecting manipulated faces can be classified into two types. One is to utilize the visual cues of the imperfections of face manipulation methods, e.g., detecting the frequency of eye blinking \cite{Blinking2018}, the abnormality of head pose \cite{Yang2019ExposingDF} and other visual features \cite{Exploiting2019}. However, these methods are usually vulnerable and can easily become invalid once the manipulation methods are refined by removing those cues. For example, Li et al. \cite{Li2018ExposingDV} put forwarded a detection method based on the face warping artifacts. This method achieved the AUC of $80.1\%$ when testing on UADFV (Published by \cite{Li2018ExposingDV}) but dropped significantly to $56.9\%$ when confronting with Celeb-DF \cite{Celeb_DF_cvpr20}. In addition to focusing on the visual cues, Li et al. \cite{Debiasi2018} introduced a detection method for morphed face images based on PRNU \cite{Chen2008, Li2012, Lin2020}.

The other category of methods resorts to the deep learning approaches. Darius et al. \cite{Afchar2018} presented two networks with a few layers to focus on the mesoscopic properties of images. Zhou et al. \cite{zhou2018twostream} proposed a method using the two-stream GoogLeNet InceptionV3 model \cite{szegedy2014going} for the task and achieved state-of-the-art performance. Hsu et al. \cite{Hsu2020} also resorted to a similar solution and used a two-stream DenseNet with contrastive loss. A triplet loss \cite{hoffer2014deep} integrated with three-stream Xception \cite{chollet2016xception} also reached state-of-the-art results, as presented by Feng et al. \cite{Feng2020}. Rössler et al. \cite{rssler2019faceforensics} also presented high performance forensics results using Xception \cite{chollet2016xception}. The robustness of the Xception \cite{chollet2016xception} network allows it to be a strong candidate for the backbone network in some other methods, e.g., the approaches introduced by Dang et al. \cite{dang2019detection} and Tolosana et al. \cite{tolosana2020deepfakes}. The rapid development of face manipulation also enables videos to be the manipulated content. Güera et al. \cite{Guera2018} used recurrent neural network (RNN) in conjunction with convolutional neural network (CNN) to detect videos containing manipulated faces. While  \cite{Guera2018} used their private dataset, Sabir et al. \cite{Sabir2019} used a similar approach to train and evaluate on FaceForensics++ \cite{rssler2019faceforensics}. 

\section{Our Approach}
\label{sec:method}

Our approach takes three steps: data preprocessing, unsupervised training, and follow-up classification. We first preprocess an image into a face-centered image to fully utilize the data of the face area. We then perform unsupervised contrastive training with the paired transformed images of each preprocessed image and the contrastive loss. This step enables the unsupervised learning of separable features, which can be used to further train a classifier for the evaluation (step 3). Figure \ref{fig:overview} shows the overview of our method.

\subsection{Data Preprocessing}
\label{subsec:dataPre}
\begin{figure*}[thbp]
\centering
\begin{minipage}[b]{1.0\linewidth}
{\label{}\includegraphics[width=1\linewidth]{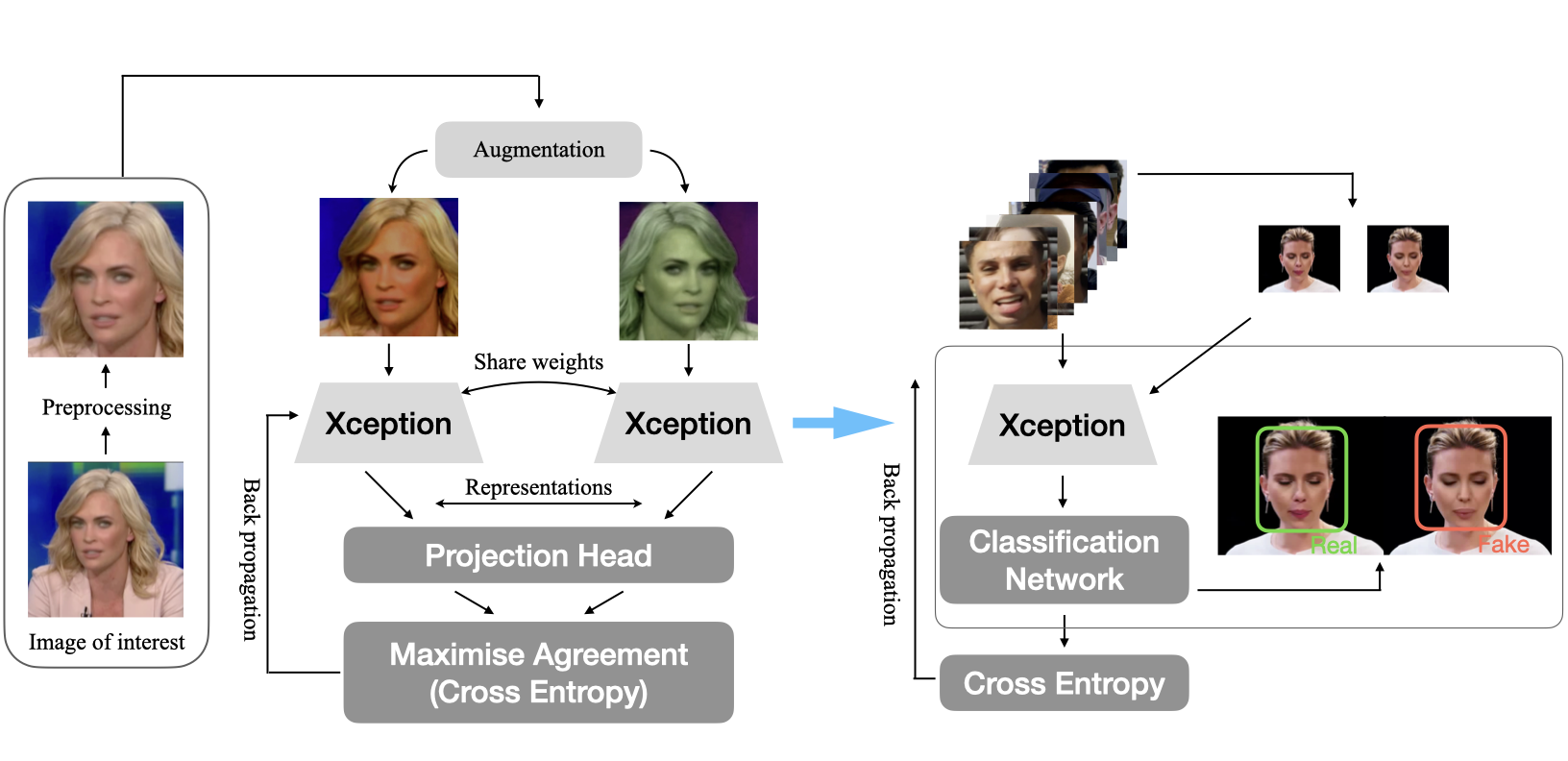}}
\end{minipage}
\caption{Overview of our proposed method. It first preprocesses an image into a face-centred image (Section \ref{subsec:dataPre}) which is then transformed into two versions using augmentation. The pair of the transformed images are further fed sequentially through the network consisting of an encoder (Xception as backbone) and projection head (a stack of linear layers) for maximizing agreement. After unsupervised learning, a simple linear classifier is trained on the features to classify images of interest as either ``real'' or ``fake''. Left: unsupervised contrastive learning, right: linear classification.    } \label{fig:overview} 
\end{figure*}

Most available face manipulation datasets are stored in the format of video. Moreover, faces in those videos usually take up a small proportion of area and thus might raise the difficulty when learning features. Therefore, following \cite{rssler2019faceforensics}, we use Dlib \cite{dlib09} for locating the bounding box of the face in each frame and crop the image with the maximum edge of the bounding box, generating a squared image with the face at the center (see Figure \ref{fig:overview}). 

\subsection{Unsupervised Contrastive Learning}
\label{subsection:networks}
At first, we transform the image of interest into its two different versions via two different augmentations (or transformations). Then we encourage the backbone network to learn the features of one image by maximizing the similarity between the two augmented versions of this image.

\begin{figure}[thbp]
\centering
\begin{minipage}[b]{1.0\linewidth}
{\label{}\includegraphics[width=1\linewidth]{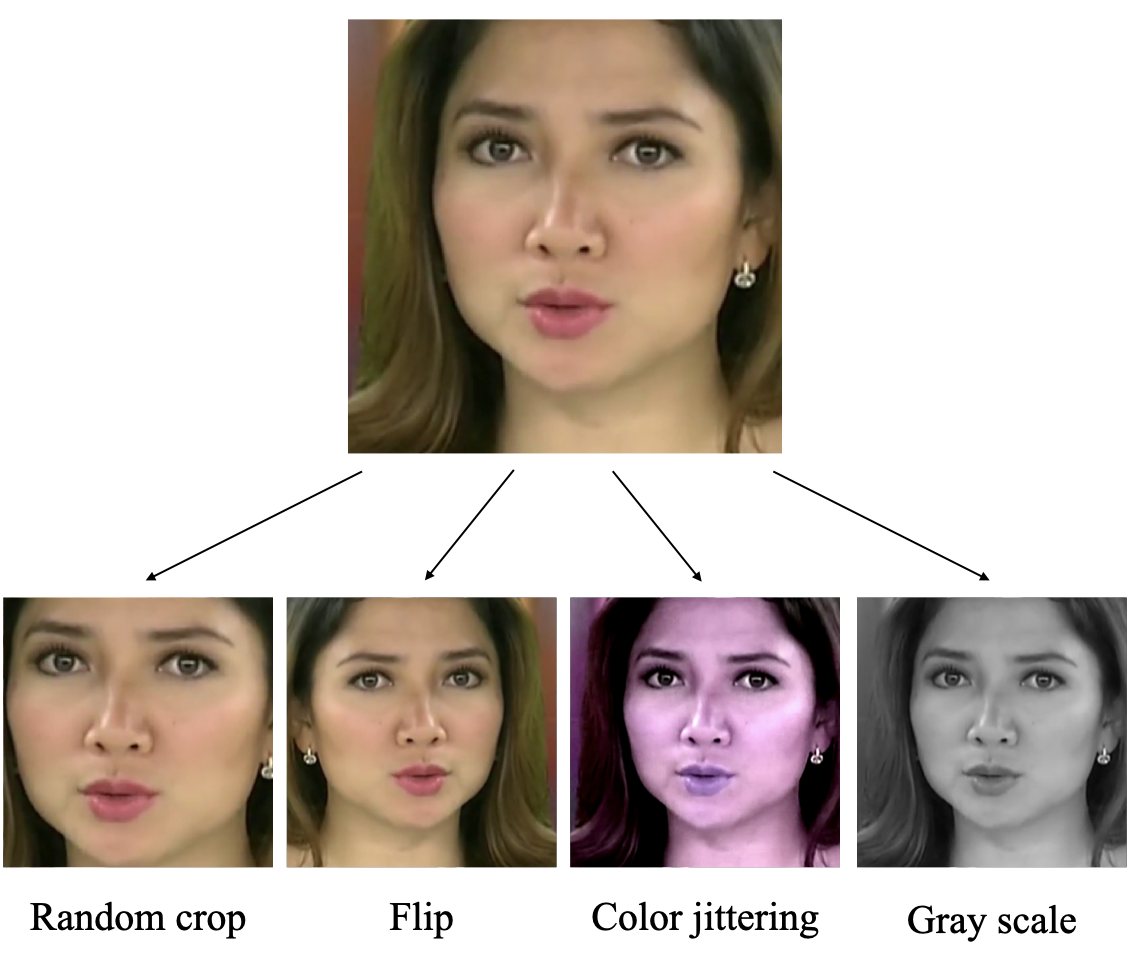}}
\end{minipage}
\caption{The basic augmentation we used when training the unsupervised learning network. Note that all augmentations are applied randomly.} \label{fig:augmentation}
\end{figure}

\textbf{Transformed versions generation.} To synthesize two different versions of each image in the dataset, we process the image of interest with several augmentation methods: random crop, random flip, random color jittering, and grayscale (Figure \ref{fig:augmentation}). 
    
We denote the image as $x$ and the two augmented views as $x_i$ and $x_j$, which will be fed into the encoder network for unsupervised contrastive learning.

\textbf{Encoder network.} Many advanced convolutional neural networks (CNN) have proven to be feasible for detecting manipulated faces in recent research, e.g., InceptionV3 \cite{zhou2018twostream}, DenseNet \cite{Hsu2020}, VGG16 \cite{dang2019detection}, Xception \cite{rssler2019faceforensics}. These CNNs are trained in a fully-supervised manner. Among these CNNs, we employ Xception as our encoder, which shows the capability of learning contrastive features \cite{Feng2020}. We denote the output of the Xception network (denote as $X_{Net}(\cdot)$) as $f_i = X_{Net}(x_i)$.

\textbf{Projection head network.} To raise the efficiency of performing the loss function, a stack of linear layers (\textit{projection head}, denoted as $g(\cdot)$) is concatenated with the Xception network. We thus obtain
\begin{equation}
    z_i = g(f_i)=W_2\sigma(W_1(f_i)),
\end{equation}
where $W_1$ and $W_2$ are two linear layers with $2,048$ and $64$ neurons, respectively. A ReLu layer $\sigma$ is placed between them. Notice that $g(\cdot)$ is only used while training the feature extraction network.

\textbf{Loss function.} We use the cosine similarity to measure the similarity of two samples.
\begin{equation}
sim(z_i, z_j) = \dfrac{z_i \cdot z_j}{\max(\Vert z_i \Vert _2 \cdot \Vert z_j \Vert _2, \epsilon)},
\end{equation}
where $\epsilon$ is set to $1e-8$. Suppose we have $N$ samples. With the pair augmentation, the mini-batch size will become $2N$. With regard to the loss, we simply apply cross-entropy after softmax regression to the similarity within a mini-batch of samples.
\begin{equation}
    Loss(i,j)= -\log \sigma_{softmax}(z_i, z_j) -\log \sigma_{softmax}(z_j, z_i),
\end{equation}
\begin{equation}
    \sigma_{softmax}(z_i, z_j) = \dfrac{exp{(\dfrac{sim(z_i, z_j)}{\tau})}}{\sum_{k=1}^{2N} exp({\dfrac{sim(z_i, z_k)}{\tau}})},
\end{equation}
Where $\tau$ is the temperature of the contrastive loss.

\subsection{Classification}
For evaluation, a variety of classifiers, such as SVM and Bayes classifier, can be trained on the features extracted by the unsupervised contrastive learning step. In this work, we simply choose a linear classification network, with the structure illustrated in Table \ref{table:classification}. It should be noted that the output of the encoder is used as input for supervised classifier training. This is because the output of the projection head involves less effective information than the output of the encoder, which is evidenced in Section \ref{Sec:ablation}.

\begin{table}[thbp]\tablefont
    \begin{center}
    \caption{The structure of the classification network.}\label{table:classification}
    \begin{tabular}{@{} c
                @{\hspace*{\lengthc}}c}
    \toprule
    Layer Number & \tabincell{l}{
    Classification network
    }\\ 
    \midrule
    1 & Fully connected layer, neurons = 2048
    \\
    2 & Fully connected layer, neurons = 4096
    \\
    3 & Fully connected layer, neurons = 2048
    \\
    4 & Fully connected layer, neurons = 256
    \\
    5 & Leaky ReLu, negative slope = 0.4
    \\
    \midrule
    6 & SoftMax Layer
    \\
    \bottomrule
    \end{tabular}
    \end{center}
\end{table}

\section{Experimental Results}
\label{sec:results}
In this section, we will first describe the used datasets and then explain the experimental settings. Then we will show the experimental results and compare the proposed method with the state-of-the-art techniques. In the end, we will provide ablation studies.

\subsection{Datasets}
We use three commonly used datasets for evaluation: FaceForensics++ \cite{rssler2019faceforensics}, UADFV \cite{Li2018ExposingDV} and Celeb-DF \cite{Celeb_DF_cvpr20}. Each dataset has its own characteristics. FaceForensics++ contains videos involving faces manipulated by a variety of methods (i.e., DeepFake \cite{Deepfakes2021}, Face2Face \cite{Face2Face2016}, FaceSwap \cite{Faceswap2021} and NeuralTexture \cite{Thies2019}). It has a total of 3,700 videos, including 1,000 pristine videos and 2,700 manipulated videos. UADFV is a relatively small yet commonly-used dataset. It consists of 98 videos, including 49 pristine videos and 49 manipulated videos. Celeb-DF is a large face forgery dataset containing faces manipulated with algorithms that are able to circumvent the common artifacts such as temporal flickering frames and color inconsistency present in the other two datasets. It contains 6,529 videos, involving 890 pristine videos and 5,639 manipulated videos. 

For each dataset, we first perform data preprocessing as illustrated in Section \ref{subsec:dataPre} and extract a maximum of 400 images for each video. Note that some videos in the dataset contain frames less than 400, and in this case, we extract all the frames for such videos. We randomly select $15\%$ of those images to form the test set and the rest are used as the training set, as suggested by Feng et al. \cite{Feng2020}. 

\begin{table}[thbp]\tablefont
    \begin{center}
    \caption{Image numbers in the split sets used in our experiments.}\label{table:amountdetail}
    \begin{tabular}{@{} l
                @{\hspace*{\lengtha}}c
                @{\hspace*{\lengtha}}c
                @{\hspace*{\lengtha}}c
                @{\hspace*{\lengtha}}c}
    \toprule
    Datasets & \tabincell{l}{
    Train (real)} & \tabincell{l}{
    Train (fake)
    } & \tabincell{l}{
    Test (real)
    } & \tabincell{l}{Test (fake)}\\ 
    \midrule
    FaceForensics++ & 115556 & 108935 & 20393 & 20473
    \\
    UADFV & 10100 & 9761 & 1783 & 1723
    \\
    Celeb-DF & 172187 & 165884 & 30386 & 29259
    \\
    \bottomrule
    \end{tabular}
    \end{center}
\end{table}

Table \ref{table:amountdetail} shows the details of the datasets we used in our experiments. Notice that the size of UADFV is less than $10\%$ of FaceForensics++ and Celeb-DF, which might affect the performance in the cross-dataset setting when the network is trained on UADFV.

\begin{figure}[thbp]
    \centering
    \begin{minipage}[b]{1.0\linewidth}
    {\label{}\includegraphics[width=1\linewidth]{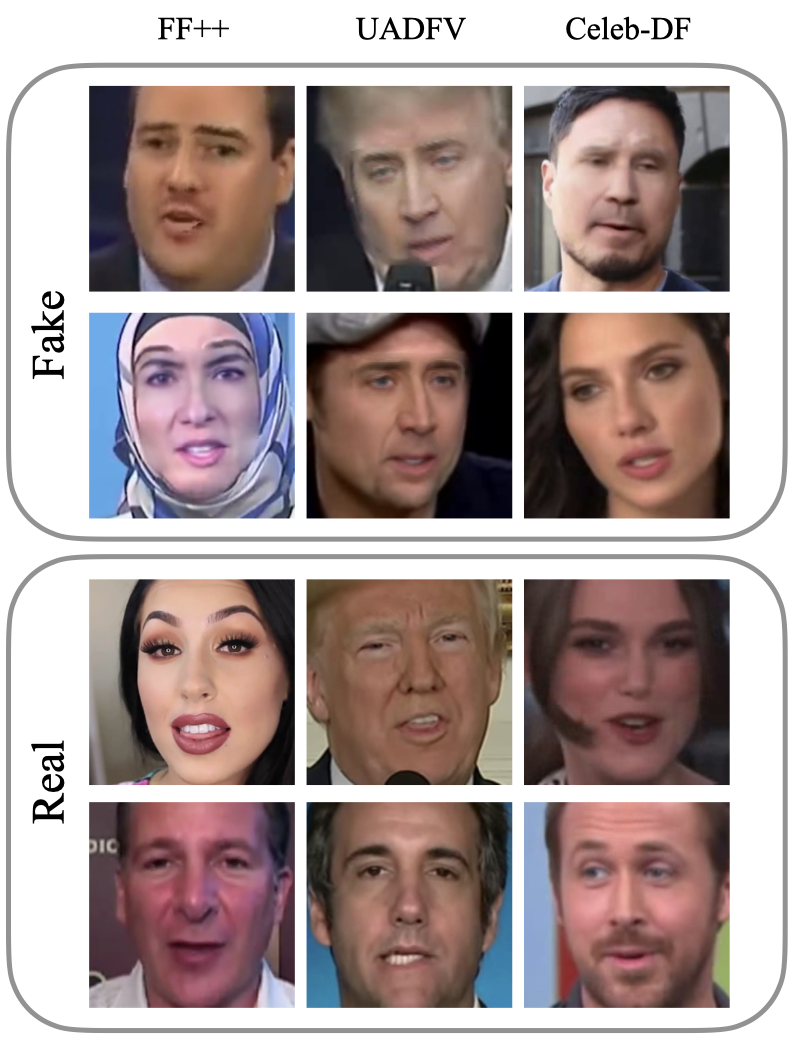}}
    \end{minipage}
    \caption{Face images after data preprocessing (face locating, cropping and resizing) from three datasets (FaceForensics++, UADFV and Celeb-DF).  }\label{fig:imageCases}
\end{figure}

We show some of the images in three datasets after preprocessing in Figure \ref{fig:imageCases}. We observe that most images from those three datasets share similar visual quality (image resolution, brightness, etc.). However, it is obvious for human eyes that the fake face images in FaceForensics++ are distinguishable because of the color inconsistency and the unnatural facial features.

\subsection{Experimental Settings}
\label{Sec:ExperimentalSetting}
Our framework is implemented on PyTorch with a desktop PC equipped with an Intel Core i9-9820X CPU (3.30GHz, 48GB memory) and a GeForce RTX 2080Ti GPU (11GB memory, CUDA 10.0). 

For training the networks, we resort to two-step learning illustrated in Figure \ref{fig:overview}. 
\begin{itemize}
\item For training the unsupervised network, the learning rate is set to $5e-4$ with the scheduler of step size 6 and 50\% descending rate. The batch size is a key factor and is set to 40 for most experiments unless otherwise specified. The temperature parameter $\tau$ is set to 0.5 throughout all experiments. 
\item For the classification task, the learning rate is set to 3e-1 with the scheduler of step size 400 and 80\% descending rate. The batch size is set to 6,000.
\end{itemize}
We employ SGD as an optimizer for both networks (unsupervised learning and classification learning) and train them for 20 epochs and 5,000 epochs, respectively.

\subsection{Comparisons}

In this section, we provide intra-dataset and cross-dataset results, which allow us to compare our method with the state-of-the-art methods.

\begin{table}[hbt!]\tablefont
    \centering
    \caption{AUC(\%) on FaceForensics++ (FF++), UADFV and Celeb-DF (See subsection \ref{subsec:dataPre}). Best results in intra-dataset setting are underlined, and best results for the cross-dataset setting are in bold.
    }\label{table:crossdatasets}
    \begin{tabular}{@{} l
                @{\hspace*{\lengthb}}c
                @{\hspace*{\lengthb}}c
                @{\hspace*{\lengthb}}c
                @{\hspace*{\lengthb}}c
                @{\hspace*{\lengthb}}c}
    \toprule
    Methods & \tabincell{l}{
    Train data
    } & \tabincell{l}{
    FF++
    } & \tabincell{l}{
    UADFV
    } & \tabincell{l}{
    Celeb-DF
    }\\ 
    \midrule
    Two-stream \cite{zhou2018twostream} & Private & 70.1 & 85.1 & 53.8
    \\
    Meso4 \cite{Afchar2018} & Private & 84.7 & 84.3& 54.8
    \\
    MesoInception4 \cite{Afchar2018} & Private & 83.0 & 82.1 & 53.6
    \\
    VA-MLP \cite{Exploiting2019}  & Private & 66.4 & 70.2 & 55.0
    \\
    VA-LogReg \cite{Exploiting2019}  & Private & 78.0 & 54.0 & 55.1
    \\
    Multi-task \cite{nguyen2019multitask}  & FF & 76.3 & 65.8 & 54.3
    \\
    Xception \cite{rssler2019faceforensics}  & FF++ & 99.7 & \textbf{80.4} & 48.2
    \\
    Capsule \cite{nguyen2019use}  & FF++ & 96.6 & 61.3 & 57.5
    \\
    Xception+Tri. \cite{Feng2020} & FF++ & \underline{99.9} & 74.3 & \textbf{61.7}
    \\
    Xception \cite{Celeb_DF_cvpr20}  & UADFV & - & 96.8 & 52.2
    \\
    Xception+Reg. \cite{dang2019detection}  & UADFV & - & 98.4 & 57.1
    \\
    Xception+Tri. \cite{Feng2020} & UADFV & 61.3 & \underline{99.9} & 60.0
    \\
    HeadPose \cite{Yang2019ExposingDF} & UADFV & 47.3 & 89.0 & 54.6
    \\
    FWA \cite{Li2018ExposingDV} & UADFV & \textbf{80.1} & 97.4 & 56.9
    \\
    Xception+Tri. \cite{Feng2020} & Celeb-DF & 60.2 & 88.9 & \underline{99.9}
    \\
    \midrule
    
    DeepfakeUCL (Ours)  & FF++ & 93.0 & 67.5 & 56.8
    \\
    DeepfakeUCL (Ours)  & UADFV & 56.2 & 98.9 & \textbf{64.8}
    \\
    DeepfakeUCL (Ours)  & Celeb-DF & 58.9 & 85.6 & 90.5
    \\
    \bottomrule
    \end{tabular}
\end{table}

\begin{figure}[thbp]
\centering
\begin{minipage}[b]{1\linewidth}
{\label{}\includegraphics[width=1\linewidth]{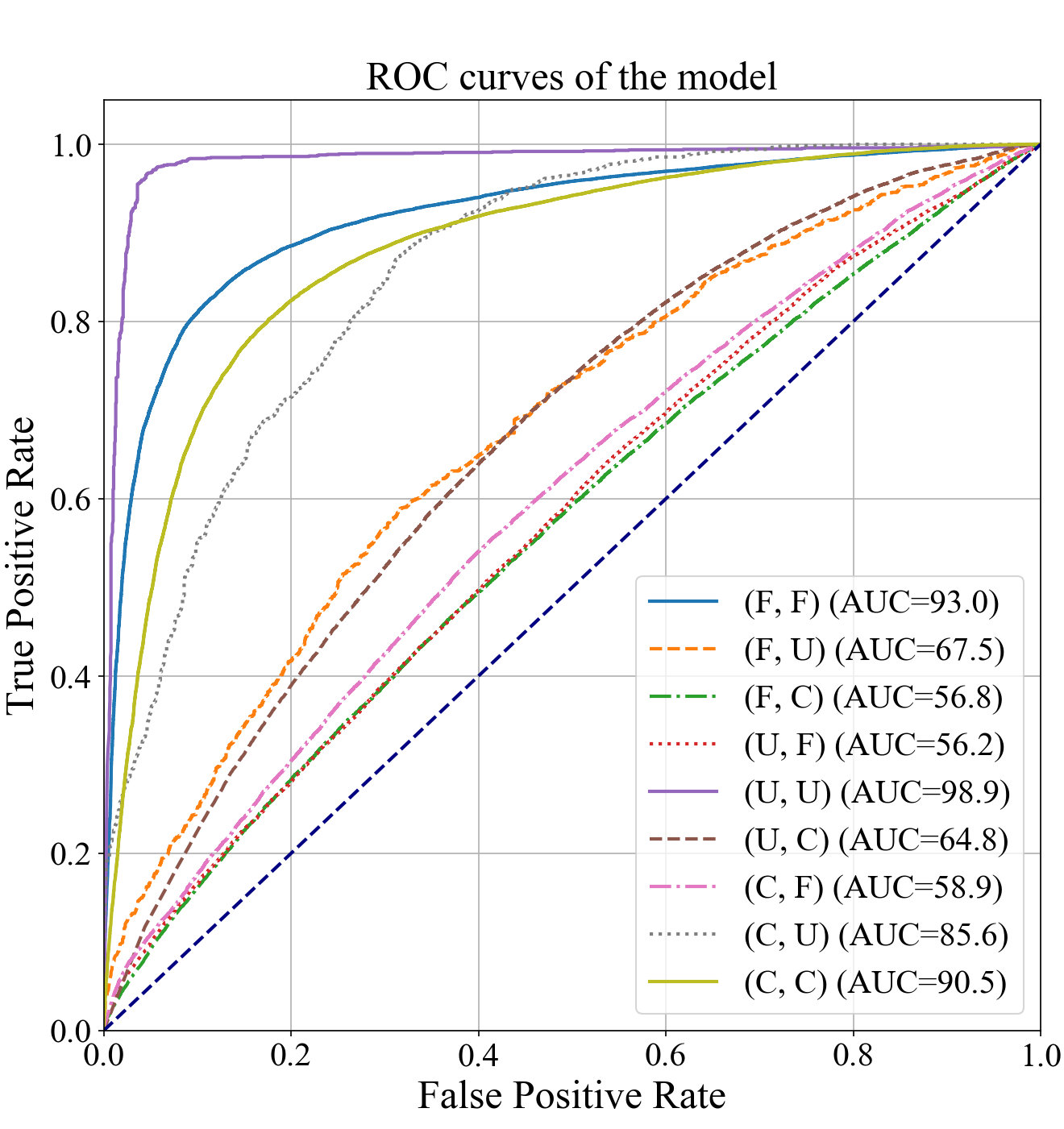}}
\end{minipage}
\caption{Receiver Operating Characteristic (ROC) curves of the models trained and tested on the same and different datasets. Note that, F, U, and C in the figure represent FaceForensics++, UADFV, and Celeb-DF, respectively. The first pair of brackets specify the datasets that the model is trained and tested on. For instance, (F, U) indicates that the model is trained on FaceForensics++ and tested on UADFV. } \label{fig:roc}
\end{figure}

We show the area under the curve (AUC) results in our experiments in Table \ref{table:crossdatasets}. Despite that our method is unsupervised, it still achieves fairly close performance and even outperforms some supervised learning methods. For results trained on FaceForensics++, our method is 3.6\% weaker than Capsule and approximately 6.8\% weaker than Xception and Xception+Tri. when testing on the same dataset. However, it outperforms Capsule and Xception by 6.2\% and 8.2\% when testing on UADFV and Celeb-DF, respectively. For results with the networks trained on UADFV, except for Xception+Tri., which is 1\% stronger, our method outperforms all other methods when testing on the same dataset. It even achieves the top results when testing on Celeb-DF, reaching 64.8\% in AUC, which is 4.8\% higher than the second-best method, Xception+Tri.. It also beats HeadPose by 8.9\% when testing on FaceForensics++, which is mere 47.3\% in AUC. When training on the Celeb-DF dataset, there are insufficient results for comparison. However, we can observe that the results of our method are quite close to Xception+Tri.. Although our method is 9.4\% weaker than Xception+Tri., it is merely 1.3\% and 3.3\% weaker than Xception+Tri. when testing on FaceForensics++ and UADFV, respectively.

We also report the ROC curves of the corresponding AUCs (in Table \ref{table:crossdatasets}) in Figure \ref{fig:roc}, where the 45-degree curve is the baseline which is achieved by a random classifier. Therefore, the closer to the baseline the curve is, the less reliable the model is. This also reveals that the inter-dataset setting is more challenging than the intra-dataset setting such that the trained models perform better at the intra-dataset setting.

\begin{figure*}[thbp]
\centering
\begin{minipage}[b]{1\linewidth}
{\label{}\includegraphics[width=1\linewidth]{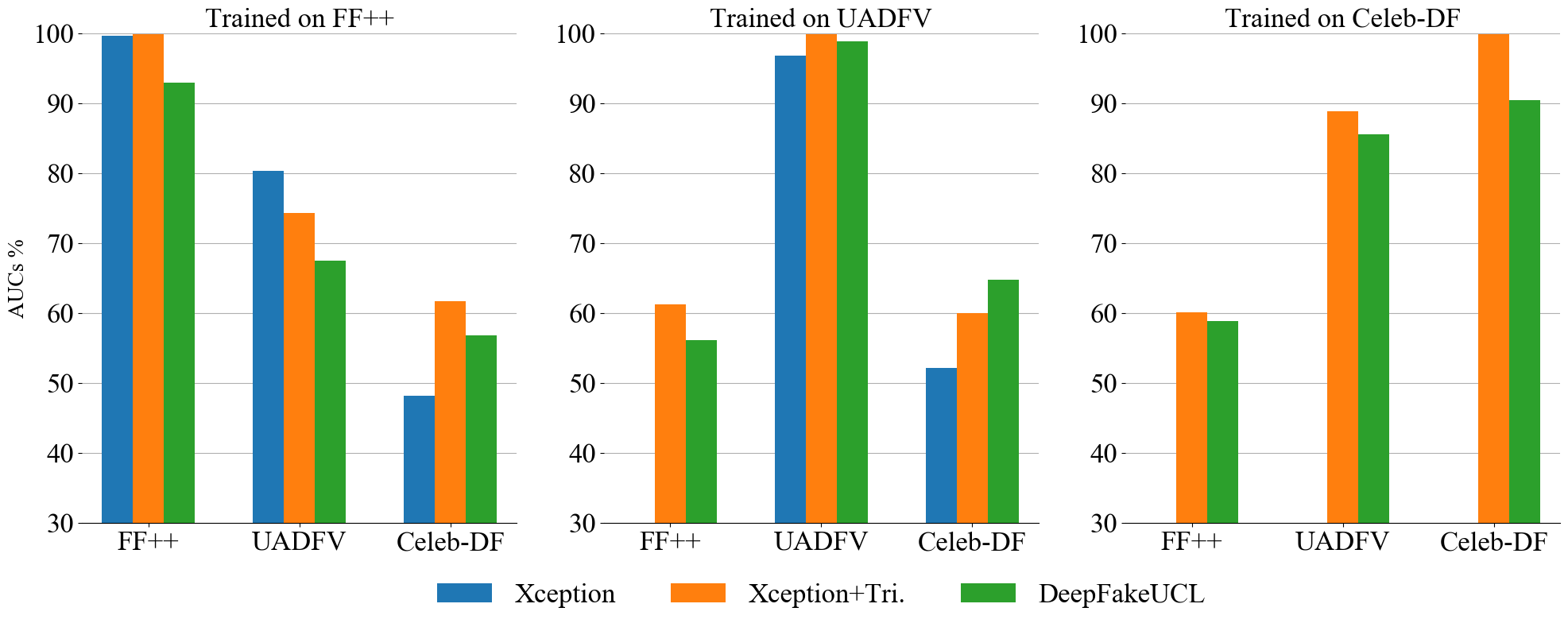}}
\end{minipage}
\caption{The data of the above three plots are based on Table \ref{table:crossdatasets}. From left to right, the networks in each plot are trained on FaceForensics++ (denoted as FF++), UADFV and Celeb-DF, respectively. In each plot, from left to right, the results in each section are tested on FaceForensics++ (also denoted as FF++), UADFV, and Celeb-DF, respectively. } \label{fig:aug_roc}
\end{figure*}

\begin{table*}[hbt!]\tablefont
    \centering
    \caption{AUC(\%) trained on Celeb-DF and tested on FaceForensics++ (FF++), UADFV and Celeb-DF with different data augmentation methods.
    }\label{table:ablation_aug}
    \begin{tabular}{@{} c
                @{\hspace*{\lengthb}}c
                @{\hspace*{\lengthb}}c
                @{\hspace*{\lengthb}}c
                @{\hspace*{\lengthb}}c
                @{\hspace*{\lengthb}}c
                @{\hspace*{\lengthb}}c
                @{\hspace*{\lengthb}}c}
    \toprule
    Augmentation 
    & \tabincell{l}{
    Random Crop
    } & \tabincell{l}{
    Random Flip
    } & \tabincell{l}{
    Random Color Jittering
    }& \tabincell{l}{
    Random Gray Scale
    }& \tabincell{l}{
    FF++
    } & \tabincell{l}{
    UADFV
    } & \tabincell{l}{
    Celeb-DF
    }\\ 
    \midrule
    1 & $\checkmark$ & $\times$ & $\times$ & $\times$ & 51.8 & 65.8  & 93.3
    \\
    2 & $\checkmark$ & $\checkmark$ & $\times$ & $\times$ & 53.8 & 64.8 & 93.2
    \\
    3 & $\checkmark$ & $\checkmark$ & $\checkmark$ & $\checkmark$ & 58.9 & 85.6 & 90.5
    \\
    \bottomrule
    \end{tabular}
\end{table*}

\subsection{Ablation Study}
\label{Sec:ablation}
In this section, we provide three ablation studies: unsupervised versus supervised contrastive learning, data augmentation and encoder versus projection head. The first study is to demonstrate if our unsupervised contrastive learning mechanism is effective. The second study is to test the effects of different combinations of data augmentation schemes. The third ablation study is to show that the output of the encoder is better than that of the projection head for the downstream classification evaluation.

\textbf{Unsupervised contrastive learning versus supervised learning.} We utilize the Xception network \cite{chollet2016xception} as the backbone of our unsupervised contrastive learning framework, which learns feature points by comparing two different views of a single image. Thus, it is useful to compare the supervised Xception networks with our method.

In Figure \ref{fig:aug_roc}, we report the comparison of three cases by taking Xception as backbone: original Xception (supervised), the combination of Xception and triplet loss (supervised), and our unsupervised contrastive learning method (DeepFakeUCL). Surprisingly, the results of our method outperform the results by training Xception on UADFV, which are 2.1\% and 12.6\% higher when testing on UADFV and Celeb-DF, respectively. Although our method is 6.7\% and 12.9\% weaker than Xception when training on FaceForensics++ and testing on FaceForensics++ and UADFV respectively, it boosts up to 56.8\% when testing on Celeb-DF, which is 8.6\% higher than the results trained by Xception. For the supervised contrastive learning method, which is driven by the Triplet network, our method still outperforms it by 4.8\%, reaching 64.8\% when training on UADFV and testing on Celeb-DF. Notice that UADFV is a relatively small dataset while Celeb-DF is about 17 times larger than UADFV (see Table \ref{table:amountdetail}). Therefore, it is considered to be a more challenging dataset when used for training purposes.

\textbf{Data augmentation choices.} To generate two different views of a single image, we utilize some common data augmentation methods to manipulate the images. However, with different data augmentation combinations, we observe some noticeably different outcomes. The results are shown in Table \ref{table:ablation_aug}. In the experiments, we utilize random image cropping as the fundamental data augmentation, which does not alter the content of the image but the position of the content in the image. Two other combinations are also used. Note that those results are all trained on Celeb-DF with the configuration illustrated in Section \ref{Sec:ExperimentalSetting}.

We can observe from the results that other than the test results of Celeb-DF itself, the AUC generally increases with respect to the complexity of the data augmentation. In other words, the more complicated the data augmentation is, the more robust the model is. When all four manipulation schemes are applied to the data augmentation process  (i.e., Augmentation 3), the results for FaceForensics++ raise dramatically to about 59\% from around 52\%. Moreover, the result for UADFV boosts enormously up to 85.6\%, which is over 20\% higher than using random cropping only. The result of testing on Celeb-DF itself drops by around 3\% when four forms of manipulation are applied, whereas this drawback seems insignificant compared to the aforementioned improvements.

\begin{figure}[thbp]
\centering
\begin{minipage}[b]{0.85\linewidth}
{\label{}\includegraphics[width=1\linewidth]{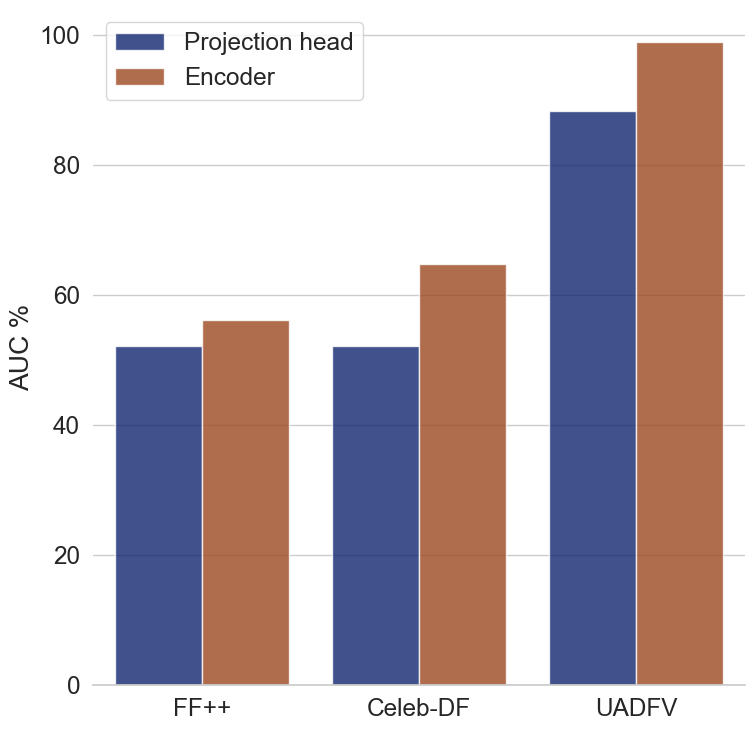}}
\end{minipage}
\caption{Encoder versus projection head. Networks are trained on the UADFV dataset.
} \label{fig:PH}
\end{figure}

\textbf{Encoder versus projection head.} We further evaluate the features learned by the projection head, by taking its output features as input for the linear classification. From Figure \ref{fig:PH}, we observe the classification results using the features output by the encoder are remarkably better than using the output of the features by the projection head, suggesting that the encoder learns more useful unsupervised features.

\section{Conclusion}
\label{sec:conclusion}
We have presented an unsupervised contrastive learning method for deepfake detection. Compared to most existing deepfake detection techniques which are fully supervised, our method learns separable features in an unsupervised manner. Experiments demonstrate the effectiveness of our method and show the comparable performance of our method to state-of-the-art deepfake detection techniques in both intra- and inter-dataset settings. We also conducted ablation studies for the proposed method. As future work, it would be interesting to incorporate temporal information within the proposed framework to achieve more robust results.

\end{document}